\documentclass[twoside,leqno,twocolumn]{article}

\usepackage[letterpaper]{geometry}

\usepackage{ltexpprt}
\usepackage{hyperref}

\usepackage{multirow}
\usepackage{booktabs}
\usepackage{amsfonts}

\usepackage{amsmath}
\usepackage{amssymb}
\usepackage{graphicx}

\usepackage{color}
\definecolor{light}{rgb}{0.3, 0.3, 0.3}
\def\light#1{{\color{light}#1}}

\usepackage{pifont}
\newcommand{\cmark}{\ding{51}}%
\newcommand{\xmark}{\ding{55}}

\begin{document}

\newcommand\relatedversion{}
\renewcommand\relatedversion{\thanks{The full version of the paper can be accessed at \protect\url{https://arxiv.org/abs/1902.09310}}} 

\title{\Large Neighbor2Seq: Deep Learning on Massive Graphs by Transforming Neighbors to Sequences}
\author{Meng Liu\thanks{Department of Computer Science and Engineering, Texas A\&M University. Email: mengliu@tamu.edu}
\and Shuiwang Ji\thanks{Department of Computer Science and Engineering, Texas A\&M University. Email: sji@tamu.edu}}

\date{}

\maketitle


\fancyfoot[R]{\scriptsize{Copyright \textcopyright\ 2022 by SIAM\\
Unauthorized reproduction of this article is prohibited}}





\begin{abstract} \small\baselineskip=9pt     Modern graph neural networks (GNNs) use a message passing scheme and
    have achieved great success in many fields. However, this recursive
    design inherently leads to excessive computation and memory
    requirements, making it not applicable to massive real-world graphs.
    In this work, we propose the Neighbor2Seq to transform the
    hierarchical neighborhood of each node into a sequence. This novel
    transformation enables the subsequent mini-batch training for general deep learning
    operations, such as convolution and attention, that are designed for
    grid-like data and are shown to be powerful in various domains. Therefore, our Neighbor2Seq naturally endows GNNs
    with the efficiency and advantages of deep learning operations on
    grid-like data by precomputing the Neighbor2Seq transformations. We evaluate our
    method on a massive graph, with more than $111$ million nodes and
    $1.6$ billion edges, as well as several medium-scale graphs. Results
    show that our proposed method is scalable to massive graphs and
    achieves superior performance across massive and medium-scale
    graphs. Our code is available at \url{https://github.com/divelab/Neighbor2Seq}.\end{abstract}


\section{Introduction}
\label{sec:intro}

Graph neural networks (GNNs) have shown effectiveness in many fields
with rich relational structures, such as citation
networks~\cite{kipf2016semi, velivckovic2017graph}, social
networks~\cite{hamilton2017inductive}, drug
discovery~\cite{gilmer2017neural, stokes2020deep, wang2020advanced}, physical
systems~\cite{battaglia2016interaction}, and point
clouds~\cite{wang2019dynamic}. Most current GNNs follow a message
passing scheme~\cite{gilmer2017neural,battaglia2018relational}, in
which the representation of each node is recursively updated by
aggregating the representations of its neighbors. Various
GNNs~\cite{li2015gated, kipf2016semi, velivckovic2017graph, xu2018powerful}
mainly differ in the forms of aggregation functions.

Real-world applications usually generate massive graphs, such as
social networks. However, message passing methods have difficulties
in handling such large graphs as the recursive message passing
mechanism leads to prohibitive computation and memory requirements.
To date, sampling methods~\cite{hamilton2017inductive,ying2018graph,chen2018stochastic,chen2018fastgcn,huang2018adaptive,zou2019layer,zeng2020graphsaint,gao2018large,chiang2019cluster,zeng2020graphsaint}
and precomputing methods~\cite{wu2019simplifying,rossi2020sign,bojchevski2020scaling}
have been proposed to scale GNNs on large graphs, as reviewed and analyzed in Section~\ref{sec:scalable_gnn}. While the sampling
methods can speed up training, they might result in redundancy,
still incur high computational complexity, lead to loss of
performance, or introduce bias.
Generally, precomputing methods can scale to larger graphs as
compared to sampling methods as recursive message passing is still
required in sampling methods.

In this work, we propose the Neighbor2Seq that transforms the
hierarchical neighborhood of each node to a sequence in a
precomputing step. After the Neighbor2Seq transformation, each node
and its associated neighborhood tree are converted to an ordered
sequence. Therefore, each node can be viewed as an independent
sample and is no longer constrained by the topological structure.
This novel transformation from graphs to grid-like data enables the
use of mini-batch training for subsequent models. As a result, our
models can be used on extremely large graphs, as long as the
Neighbor2Seq step can be precomputed.

As a radical departure from existing precomputing methods, including SGC~\cite{wu2019simplifying} and SIGN~\cite{rossi2020sign}, we
consider the hierarchical neighborhood of each node as an ordered
sequence. This order information corresponding to hops between nodes is preserved in our Neighbor2Seq and captured explicitly in subsequent order sensitive operations, such as convolutions. This order information is vital for improving the prediction, as confirmed by our experiments and ablation studies in Section~\ref{sec:exp}.
As a result of our Neighbor2Seq transformation, generic and powerful deep
learning operations for grid-like data, such as convolution and
attention, can be applied in subsequent models to learn from the resulting sequences. Experimental results indicate that our proposed method can be used
on a massive graph, where most current methods cannot be applied.
Furthermore, our method achieves superior performance as compared
with previous sampling and precomputing methods.

\section{Analysis of Current Graph Neural Network Methods}\label{sec:mpnn}

We start by defining necessary notations. A graph is formally defined as
$\mathcal{G} = (V, E)$, where $V$ is the set of nodes and $E
\subseteq V \times V$ is the set of edges. We use $n=|V|$ and
$m=|E|$ to denote the numbers of nodes and edges, respectively. The
nodes are indexed from $1$ to $n$. We consider a node feature matrix
$\boldsymbol{X} \in \mathbb{R}^{n \times d}$, where each row
$\boldsymbol{x}_i \in \mathbb{R}^d$ is the $d$-dimensional feature
vector associated with node $i$. The topology information of the
graph is encoded in the adjacency matrix $\boldsymbol{A} \in
\mathbb{R}^{n \times n}$, where $\boldsymbol{A}_{(i, j)} = 1$ if an edge exists
between node $i$ and node $j$, and $\boldsymbol{A}_{(i, j)} = 0$ otherwise. Although we describe our approach on undirected graphs for simplicity, it can be naturally applied on directed graphs.

\subsection{Graph Neural Networks via Message Passing.}\label{sec:gnn}

There are two primary deep learning methods on
graphs~\cite{bronstein2017geometric}; those are, spectral methods~\cite{bruna2013spectral}
and spatial methods. In this work, we focus on
the analysis of the current mainstream spatial methods. Generally,
most existing spatial methods, such as
ChebNet~\cite{defferrard2016convolutional},
GCN~\cite{kipf2016semi}, GG-NN~\cite{li2015gated},
GAT~\cite{velivckovic2017graph}, and GIN~\cite{xu2018powerful},
can be understood from the message passing
perspective~\cite{gilmer2017neural, battaglia2018relational}.
Specifically, we iteratively update node representations by
aggregating representations from its immediate neighbors. These
message passing methods have been shown to be effective in many
fields. However, they have inherent difficulties when applied on
large graphs due to their excessive computation and memory
requirements, as described in Section~\ref{sec:scalable_gnn}.

\subsection{Graph Neural Networks on Large Graphs.}\label{sec:scalable_gnn}


The above message passing methods are often trained in full batch.
This requires the whole graph, \emph{i.e.}, all the node
representations and edge connections, to be in memory to allow
recursive message passing on the whole graph. Usually, the number of
neighbors grows very rapidly with the increase of receptive field~\cite{alon2020bottleneck}.
Hence, these methods cannot be applied directly on large-scale
graphs due to the prohibitive computation and
memory. To enable deep learning on large graphs, two families of
methods have been proposed; those are methods based on sampling and
precomputing.


To circumvent the recursive expansion of neighbors across layers,
sampling methods apply GNNs on a sampled subset of nodes with
mini-batch training. Sampling methods can be further divided
into three categories. First, node-wise sampling methods perform
message passing for each node in its sampled neighborhood. This
strategy is first proposed in
GraphSAGE~\cite{hamilton2017inductive}, where neighbors are
randomly sampled. This is extended by PinSAGE~\cite{ying2018graph},
which selects neighbors based on random walks.
VR-GCN~\cite{chen2018stochastic} further proposes to use variance
reduction techniques to obtain a convergence guarantee. Although
these node-wise sampling methods can reduce computation, the
remaining computation is still very expensive and some redundancy
might have been introduced, as described
in~\cite{huang2018adaptive}. Second, layer-wise sampling methods
sample a fixed number of nodes for each layer of GNNs. In particular,
FastGCN~\cite{chen2018fastgcn} samples a fixed number of nodes for
each layer independently based on the degree of each node.
AS-GCN~\cite{huang2018adaptive} and LADIES~\cite{zou2019layer}
introduce between-layer dependencies during sampling, thus
alleviating the loss of information. Layer-wise sampling methods can
avoid the redundancy introduced by node-wise sampling methods.
However, the expensive sampling algorithms that aim to ensure
performance may themselves incur high computational cost, as pointed
out in~\cite{zeng2020graphsaint}. Third, graph-wise sampling methods
build mini-batches on sampled subgraphs. Specifically,
LGCN~\cite{gao2018large} proposes to leverage mini-batch training
on subgraphs selected by Breadth-First-Search algorithms.
ClusterGCN~\cite{chiang2019cluster} conducts mini-batch training
on sampled subgraphs that are obtained by a graph clustering
algorithm. GraphSAINT~\cite{zeng2020graphsaint} proposes to derive
subgraphs by importance-sampling and introduces normalization
techniques to eliminate biases. These graph-wise sampling methods
usually have high efficiency. The main limitation is that the nodes
in a sampled subgraph are usually clustered together. This implies
that two distant nodes in the original graph usually cannot be
feeded into the GNNs in the same mini-batch during training,
potentially leading to bias in the trained models.

\begin{figure*}[t]
    \begin{center}
    \includegraphics[width=0.8\textwidth]{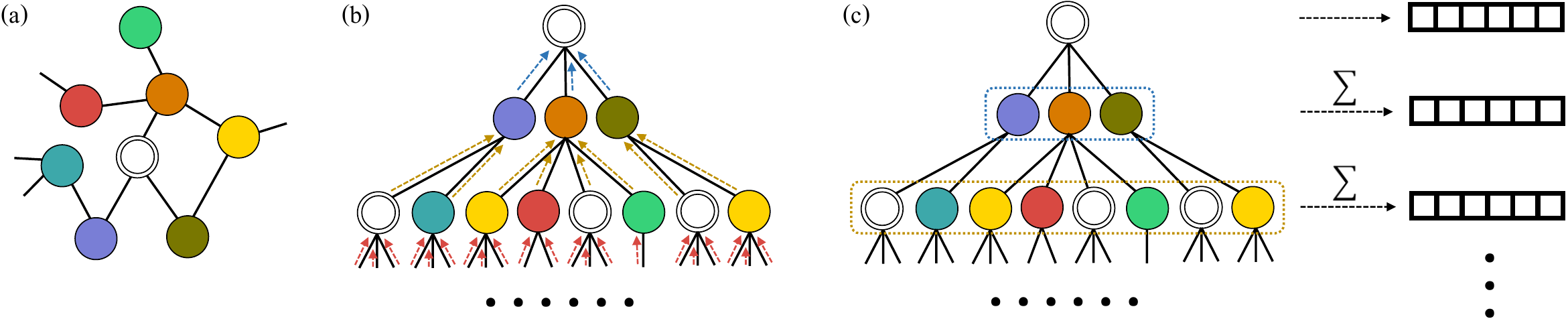}
    \end{center}%
    \vspace{-0.6cm}
    \caption{(a) An illustration of the original graph. The current node is denoted as two concentric circles. (b) Message passing in the neighborhood tree. (c) Our proposed Neighbor2Seq.}
    \label{fig:neighborhood_tree}
\end{figure*}


The second family of methods for enabling GNNs training on large
graphs are based on precomputing. Specifically,
SGC~\cite{wu2019simplifying} removes the non-linearity between GCN
layers, resulting in a simplification as $\boldsymbol{Y} =
\textrm{softmax}(\boldsymbol{\hat{A}}^{L}\boldsymbol{X}\boldsymbol{W})$.
In this formulation, $\boldsymbol{\hat{A}} =
\boldsymbol{\tilde{D}}^{-\frac{1}{2}}\boldsymbol{\tilde{A}}\boldsymbol{\tilde{D}}^{-\frac{1}{2}}$
is the symmetrically normalized adjacency matrix,
$\boldsymbol{\tilde{A}} = \boldsymbol{A} + \boldsymbol{I}$ is the
adjacency matrix with self-loops, $\boldsymbol{\tilde{D}}$ is the
corresponding diagonal node degree matrix with
$\boldsymbol{\tilde{D}}_{(i,i)} = \sum_j
\boldsymbol{\tilde{A}}_{(i,j)}$, $L$ is the size of receptive field
(\emph{i.e.}, the number of considered neighboring hops), which is
the same as an $L$-layer GCN, $\boldsymbol{Y}$ is the output of the
softmax classifier. Since there is no learnable parameters in
$\boldsymbol{\hat{A}}^{L}\boldsymbol{X}$, this term can be
precomputed as a feature pre-processing step. Similarly,
SIGN~\cite{rossi2020sign} applies an inception-like model to the
precomputed features $\boldsymbol{\hat{A}}^{\ell}\boldsymbol{X}$ for
$\ell \in \{1, \cdots, L\}$, where $L$ is the predefined size of
receptive field. Instead of precomputing the smoothing features as
in SGC and SIGN, PPRGo~\cite{bojchevski2020scaling} extends the
idea of PPNP~\cite{klicpera2018predict} by approximately
precomputing the personalized PageRank~\cite{page1999pagerank}
matrix, thereby enabling model training on large graphs using
mini-batches. Generally, the precomputing methods can scale to
larger graphs because the sampling methods still need
to perform the recursive message passing during training. Differing
from these precomputing methods, we consider the hierarchical
neighborhood of each node as an ordered sequence, thus retaining the
useful information about hops between nodes and enabling subsequent
powerful and efficient operations.

\section{The Proposed Neighbor2Seq Method and Analysis}\label{sec:method}
In this section, we firstly describe our proposed method, known as
Neighbor2Seq, which transforms the hierarchical neighborhood of each
node into an ordered sequence, thus enabling the subsequent use of
general deep learning operations. Then, we analyze the scalability of our
method (Section~\ref{sec:scalablitiy}).

\subsection{Overview.}


As described in Section~\ref{sec:gnn}, existing message passing
methods recursively update each node's representation by aggregating
information from its immediate neighbors. Hence, what these methods
aim at capturing for each node is essentially its corresponding
hierarchical neighborhood, \emph{i.e.}, the neighborhood tree rooted
at the current node, as illustrated in Figure~\ref{fig:neighborhood_tree} (b). In
this work, we attempt to go beyond the message passing scheme to
overcome the limitations mentioned in Section~\ref{sec:scalable_gnn}. We
propose to capture the information of this hierarchical neighborhood
by transforming it into an ordered sequence, instead of recursively
squashing it into a fixed-length vector. Our proposed method is
composed of three steps. First, we transform the neighborhood to a
sequence for each node. Second, we apply a normalization technique
to the derived sequence features. Third, we use general deep
learning operations, \emph{i.e.}, convolution and attention, to
learn from these sequence features and then make predictions for
nodes. In the following, we describe these three steps in detail.

\subsection{Neighbor2Seq: Transforming Neighborhoods to Sequences.}

The basic idea of the proposed Neighbor2Seq is to transform the hierarchical
neighborhood of each node to an ordered sequence by integrating the
features of nodes in each layer of the neighborhood tree. Following
the notations defined in Section~\ref{sec:mpnn}, we let
$\boldsymbol{z}^i_0, \boldsymbol{z}^i_1, \cdots, \boldsymbol{z}^i_L$
denote the resulting sequence for node $i$, where $L$ is the height of the neighborhood
tree rooted at node $i$, \emph{i.e.}, the number of hops we consider. $\boldsymbol{z}^i_\ell \in \mathbb{R}^d$
denotes the $\ell$-th feature of the sequence. The length of the
resulting sequence for each node is $L+1$. Formally, for each node
$i$, our Neighbor2Seq can be expressed as
\begin{equation}\label{eq:neighbor2seq}
\boldsymbol{z}^i_\ell = \sum_{j=1}^{n} w(i, j,
\ell)\boldsymbol{x}_j, \quad \forall \ell \in \{0, 1, 2, \cdots,
L\},
\end{equation}
where $w(i, j, \ell)$ denotes the number of walks with length $\ell$
between node $i$ and $j$. $n$ is the number of nodes in the graph.
We define $w(i, j, 0)$ as $1$ for $j=i$ and $0$ otherwise. Hence,
$\boldsymbol{z}^i_0$ is the original node feature
$\boldsymbol{x}_i$. Intuitively, $\boldsymbol{z}^i_\ell$ is obtained
by computing a weighted sum of features of all nodes with walks of
length $\ell$ to $i$, and the numbers of qualified walks are used as
the corresponding weights. Our Neighbor2Seq is illustrated in
Figure~\ref{fig:neighborhood_tree} (c). Note that the derived sequence has
meaningful order information, indicating the hops between nodes.
After we obtain ordered sequences from the original hierarchical
neighborhoods, we can use generic deep learning operations to learn
from these sequences, as detailed below.

\begin{figure*}[t]
    \begin{center}
        \includegraphics[width=0.8\textwidth]{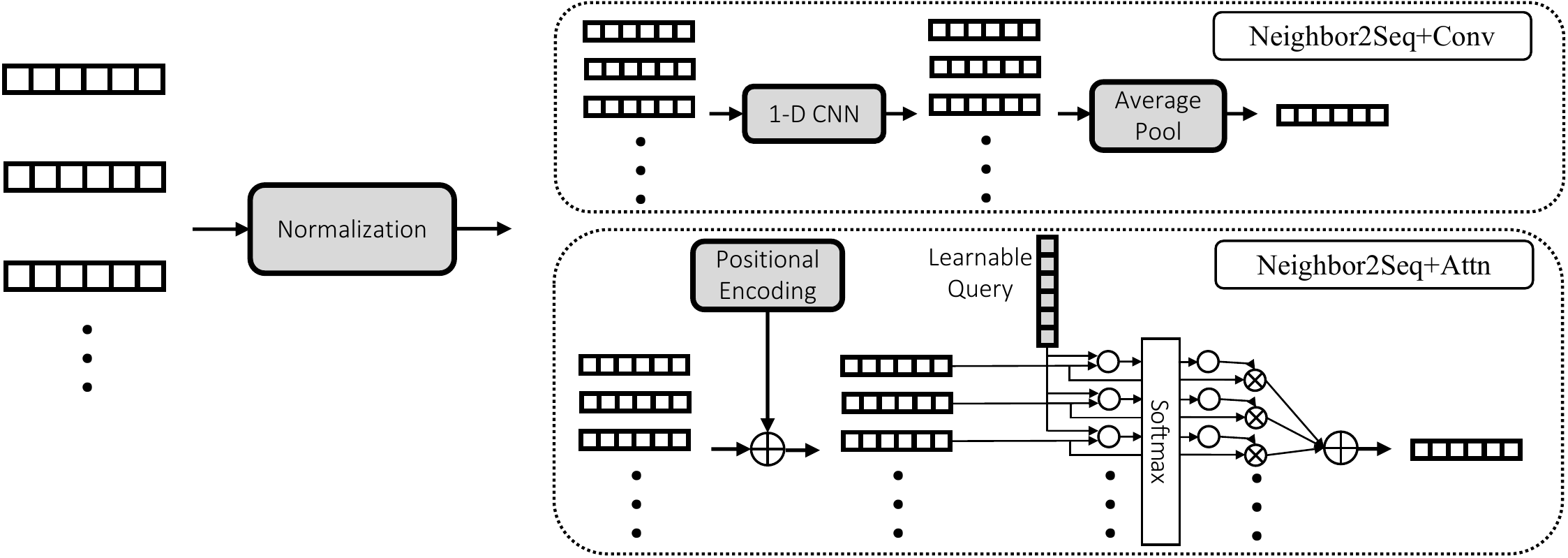}
    \end{center}%
    \vspace{-0.6cm}
    \caption{An illustration of our proposed models: Neighbor2Seq+Conv and Neighbor2Seq+Attn.}
    \label{fig:conv_attn}
\end{figure*}

\subsection{Normalization.}

Since the number of nodes in the hierarchical neighborhood grows
exponentially as the hop number increases, different layers in the
neighborhood tree have drastically different numbers of nodes.
Hence, feature vectors of a sequence computed by
Eq.~(\ref{eq:neighbor2seq}) have very different scales. In
order to make the subsequent learning easier, we propose a layer to
normalize the sequence features. We use a normalization technique
similar to layer normalization~\cite{ba2016layer}. In particular,
each feature of a sequence is normalized based on the mean and the
standard deviation of its own feature values. Formally, our
normalization process for each node $i$ can be written as
\begin{equation}
\begin{aligned}
\boldsymbol{y}^i_\ell &=
\boldsymbol{W_\ell}\boldsymbol{z}^i_\ell, &\quad\forall
\ell \in \{0, 1, 2, \cdots, L\},\\
\boldsymbol{o}^i_\ell
&= \frac{\boldsymbol{y}^i_\ell -
    \mu^i_\ell}{\sigma^i_\ell}\odot\boldsymbol{\gamma}_\ell+\boldsymbol{\beta}_\ell,
&\quad\forall
\ell \in \{0, 1, 2, \cdots, L\},
\end{aligned}
\end{equation}
where $\mu^i_\ell$ and $\sigma^i_\ell$ denote the corresponding mean and standard deviation
of the representation $\boldsymbol{y}^i_\ell$. Formally,
\begin{equation}
\begin{aligned}
&  \mu^i_\ell=\frac{1}{d'}\sum_{c=1}^{d'}\boldsymbol{y}^i_\ell[c],
\quad
\sigma^i_\ell=\sqrt{\frac{1}{d'}\sum_{c=1}^{d'}(\boldsymbol{y}^i_\ell[c]-\mu^i_\ell)^2}.
\end{aligned}
\end{equation}
We first apply a learnable linear transformation $\boldsymbol{W_\ell} \in
\mathbb{R}^{d' \times d}$ to produce a low-dimensional
representation $\boldsymbol{y}^i_\ell \in \mathbb{R}^{d'}$ for the
$\ell$-th feature of the sequence, since the original feature
dimension $d$ is usually large.
$\boldsymbol{\gamma}_\ell \in \mathbb{R}^{d'}$ and
$\boldsymbol{\beta}_\ell \in \mathbb{R}^{d'}$ denote the learnable
affine transformation parameters. $\odot$ denotes the element-wise
multiplication. Note that the learnable parameters in this
normalization layer is associated with $\ell$, implying that each
feature of the sequence is normalized separately. Using this
normalization layer, we obtain the normalized feature vector
$\boldsymbol{o}^i_\ell \in \mathbb{R}^{d'}$ for every $ \ell \in
\{0, 1, 2, \cdots, L\}$.

\subsection{Neighbor2Seq+Conv and Neighbor2Seq+Attn.}

After obtaining an ordered sequence for each node, we can view each
node in the graph as a sequence of feature vectors. We can use
general deep learning techniques to learn from these sequences. In
this work, we propose two models, namely Neighbor2Seq+Conv and
Neighbor2Seq+Attn, in which convolution and attention are applied to
the sequences of each node accordingly.

As illustrated in Figure~\ref{fig:conv_attn}, Neighbor2Seq+Conv
applies a 1-D convolutional neural network to the sequence features
and then use an average pooling to yield a representation for the
sequence. Formally, for each node $i$,
\begin{equation}
\begin{aligned}
\left(\boldsymbol{\hat{o}}^i_0, \boldsymbol{\hat{o}}^i_1, \cdots,
\boldsymbol{\hat{o}}^i_L\right) &=
\textrm{CNN}\left(\boldsymbol{o}^i_0, \boldsymbol{o}^i_1, \cdots,
\boldsymbol{o}^i_L\right),\\
\boldsymbol{r}^i &=
\frac{1}{L+1} \sum_{\ell=0}^{L} \boldsymbol{\hat{o}}^i_\ell,
\end{aligned}
\end{equation}
where $\textrm{CNN}(\cdot)$ denotes a 1-D convolutional neural
network. $\boldsymbol{r}^i$ denotes the obtained representation of
node $i$ that is used as the input to a linear classifier to make
prediction for this node. Specifically, we implement
$\textrm{CNN}(\cdot)$ as a 2-layer convolutional neural network
composed of two 1-D convolutions. The activation function between
layers is ReLU~\cite{krizhevsky2012imagenet}.

Incorporating attention is another natural idea to learn from
sequences. As shown in Figure~\ref{fig:conv_attn},
Neighbor2Seq+Attn uses an attention
mechanism~\cite{bahdanau2015neural} to integrate sequential feature
vectors in order to derive an informative representation for each node. Unlike convolutional
neural networks, the vanilla attention mechanism cannot make use of
the order of the sequence. Hence, we add positional
encodings~\cite{vaswani2017attention} to the features such that the
position information of different features in the sequence can be
incorporated. Formally, for each node $i$, we add positional
encoding for each feature in the sequence as
\begin{equation}
\begin{aligned}
\boldsymbol{p}^i_\ell[m] &= \left\{
\begin{aligned}
&\sin \left(\frac{\ell}{10000^{\frac{2n}{d'}}}\right)\quad m=2n&\\
&\cos \left(\frac{\ell}{10000^{\frac{2n}{d'}}}\right)\quad m=2n+1&
\end{aligned},
\right . \\ 
\boldsymbol{k}^i_\ell &= \boldsymbol{o}^i_\ell +
\boldsymbol{p}^i_\ell. 
\end{aligned}
\end{equation}
The positional encoding for $\ell$-th feature of node $i$ is denoted
as $\boldsymbol{p}^i_\ell \in \mathbb{R}^{d'}$. $m \in \{1, 2,
\cdots, d'\}$ is the dimensional index. Intuitively, a
position-dependent vector is added to each feature such that the
order information can be captured. Then we use the attention
mechanism with learnable query~\cite{yang2016hierarchical} to
combine these sequential feature vectors to obtain the final
representations $\boldsymbol{r}^i$ for each node $i$. Formally,
\begin{equation}
\boldsymbol{r}^i = \sum_{\ell=0}^{L} \alpha^i_\ell
\boldsymbol{k}^i_\ell, \quad \alpha^i_\ell  =
\frac{\exp({\boldsymbol{k}^i_\ell}^T\boldsymbol{q})}{\sum_{\ell=0}^{L}
    \exp({\boldsymbol{k}^i_\ell}^T\boldsymbol{q})}.
\end{equation}
$\boldsymbol{q} \in \mathbb{R}^{d'}$ is the learnable query vector
that is trained along with other model parameters. The derived
representation $\boldsymbol{r}^i$ will be taken as the input to a
linear classifier to make prediction for node $i$.

\subsection{Analysis of Scalability.}\label{sec:scalablitiy}

\textit{(i) Precomputing Neighbor2Seq.} A well-known fact is that the
value of $w(i, j, \ell)$ in Eq.~(\ref{eq:neighbor2seq}) can be
obtained by computing the power of the original adjacency matrix
$\boldsymbol{A}$. Following GCN, we add self-loops to make each node
connect to itself. Concretely, $w(i, j, \ell) =
\boldsymbol{\tilde{A}}^\ell_{(i, j)}$, where $\boldsymbol{\tilde{A}}=\boldsymbol{A}+\boldsymbol{I}$. Hence, the Neighbor2Seq can be
implemented by computing the matrix multiplications
$\boldsymbol{\tilde{A}}^\ell\boldsymbol{X}$ for $\forall \ell \in \{0, 1,
2, \cdots, L\}$. Note that we do not need to normalize the adjacency matrix in our Neighbor2Seq, differing from popular GNNs, such as GCN, and existing precomputing methods, such as SGC and SIGN. Since there is no learnable parameters in the
Neighbor2Seq step, these matrix multiplications can be precomputed
sequentially for large graphs on CPU platforms with large memory.
This can be easily precomputed because the matrix
$\boldsymbol{\tilde{A}}$ is usually sparse. For extremely large
graphs, this precomputation can even be performed on distributed
systems.

\textit{(ii) Enabling mini-batch training.} After we obtain the
precomputed sequence features, each node in the graph corresponds to
a sequence of feature vectors. Therefore, each node can be viewed as
an independent sample. That is, we are no longer restricted by the
original graph connectivity. Then, we can randomly sample
from all the training nodes to conduct mini-batch training. This is
more flexible and unbiased than sampling methods as reviewed
in Section~\ref{sec:scalable_gnn}. Our mini-batches can be randomly
extracted over all nodes, opening the possibility that any pair of
nodes can be sampled in the same mini-batch. In contrast,
mini-batches in sampling methods are usually restricted by the
fixed sampling strategies. This advantage opens the door for
subsequent model training on extremely large graphs, as long as the
corresponding Neighbor2Seq step can be precomputed.

\begin{table}[t]
    \caption{Comparison of computational complexity for precomputing and forward pass (per epoch).}
    \label{tab:computation}
    \centering
    \vspace{0.05in}
    \resizebox{1\columnwidth}{!}{
        \begin{tabular}{lcc}
            \toprule
            Method & Precomputing & Forward Pass \\
            \midrule
            GCN & - & $\mathcal{O}(Ldm+Ld^2n)$ \\
            GraphSAGE & $\mathcal{O}(s^Ln)$ & $\mathcal{O}(s^Ld^2n)$ \\
            ClusterGCN & $\mathcal{O}(m)$  & $\mathcal{O}(Ldm+Ld^2n)$ \\
            GraphSAINT & $\mathcal{O}(sn)$ & $\mathcal{O}(Ldm+Ld^2n)$ \\
            SGC & $\mathcal{O}(Ldm)$ & $\mathcal{O}(d^2n)$ \\
            SIGN & $\mathcal{O}(Ldm)$ & $\mathcal{O}(Ld^2n)$ \\
            \midrule
            Neighbor2Seq+Conv & $\mathcal{O}(Ldm)$ & $\mathcal{O}((Ld^2+Lkd^2)n)$ \\
            Neighbor2Seq+Attn & $\mathcal{O}(Ldm)$ & $\mathcal{O}((Ld^2+Ld)n)$ \\
            \bottomrule
        \end{tabular}
        }
\end{table}

\begin{table*}[t]
    \caption{Statistics of datasets. ``m'' denotes multi-label classification.}
    \label{tab:datsets}
    \centering
    \vspace{0.05in}
        \resizebox{2\columnwidth}{!}{
            \begin{tabular}{lcccccccc}
                \toprule
                \textbf{Dataset}& \textbf{Scale} &\textbf{Task} & \textbf{\#Nodes} & \textbf{\#Edges}  & \textbf{Avg. Deg.} & \textbf{\#Features} & \textbf{\#Classes} & \textbf{Train/Val/Test} \\
                \midrule
                \textit{ogbn-papers100M} & Massive & Transductive &$111, 059, 956$ & $1,615,685,872$ & $29$ & $128$ & $172$ & $0.78/0.08/0.14$ \\
                \textit{ogbn-products} & Medium & Transductive &$2, 449, 029$ & $61, 859, 140$ & $51$ & $100$ & $47$ & $0.08/0.02/0.90$ \\
                \textit{Reddit} & Medium & Inductive & $232, 965$ & $11, 606, 919$ & $50$ & $602$ & $41$ & $0.66/0.10/0.24$ \\
                \textit{Yelp} & Medium & Inductive & $716, 857$ & $6, 997, 410$ & $10$ & $300$ & $100$(m) & $0.75/0.10/0.15$ \\
                \textit{Flickr} & Medium & Inductive & $89, 250$ & $899, 756$ & $10$ & $500$ & $7$ & $0.50/0.25/0.25$ \\
                \bottomrule
            \end{tabular}
        }
\end{table*}

\begin{table}[t]
    \caption{Results on \textit{ogbn-papers100M} in terms of classification accuracy (in percent). The reported accuracy is averaged over $10$ random runs. The best performance on test set is highlighted in bold and the second performance is shown with an underline.}
    \label{tab:result_papers100M}
    \centering
    \vspace{0.05in}
    \resizebox{\columnwidth}{!}{
        \begin{tabular}{lccc}
            \toprule
            Method & Training & Validation & Test\\
            \midrule
            MLP & $54.84 \scriptstyle{{\light{\pm0.43}}}$ & $49.60\scriptstyle{{\light{\pm0.29}}}$ & $47.24\scriptstyle{{\light{\pm0.31}}}$ \\
            Node2vec & - & $58.07\scriptstyle{{\light{\pm0.28}}}$ & $55.60\scriptstyle{{\light{\pm0.23}}}$ \\
            SGC & $67.54\scriptstyle{{\light{\pm0.43}}}$ & $66.48\scriptstyle{{\light{\pm0.20}}}$ & $63.29\scriptstyle{{\light{\pm0.19}}}$ \\
            \midrule
            \textbf{Neighbor2Seq+Conv} & $70.94\scriptstyle{{\light{\pm1.40}}}$ & $68.19\scriptstyle{{\light{\pm0.32}}}$ & $\textbf{65.62}\scriptstyle{{\light{\pm0.32}}}$ \\
            \textbf{Neighbor2Seq+Attn} & $70.75\scriptstyle{{\light{\pm0.68}}}$ & $67.60\scriptstyle{{\light{\pm0.31}}}$ & $\underline{65.04}\scriptstyle{{\light{\pm0.24}}}$ \\
            \bottomrule
        \end{tabular}
    }
\end{table}

\textit{(iii) Computational complexity comparison.} We compare our methods
with several existing sampling and precomputing methods in terms of
computational complexity. We let $L$ denote the number of hops we
consider. For simplicity, we assume the feature dimension $d$ is
fixed for all layers. For sampling methods, $s$ is the number of
sampled neighbors for each node. The computation of
Neighbor2Seq+Conv mainly lies in the linear transformation
(\emph{i.e.}, $\mathcal{O}(Ld^2n)$) in the normalization step and
the 1-D convolutional neural networks (\emph{i.e.},
$\mathcal{O}(Lkd^2n)$, where $k$ is the kernel size). Hence, the
computational complexity for the forward pass of Neighbor2Seq+Conv
is $\mathcal{O}((Ld^2+Lkd^2)n)$. Neighbor2Seq+Attn has a
computational complexity of $\mathcal{O}((Ld^2+Ld)n)$ because the
attention mechanism using learnable query is more efficient than 1-D convolutional neural
networks. As shown in Table~\ref{tab:computation}, the forward pass
complexities of precomputing methods, including our
Neighbor2Seq+Conv and Neighbor2Seq+Attn, are all linear with respect
to the number of nodes $n$ and do not depend on the number of edges
$m$. Hence, the training processes of our models are computationally
efficient. Compared with existing precomputing methods, our models, especially Neighbor2Seq+Conv, are slightly more time consuming due to the introduced convolution or attention operations. However, our models have the same order of time complexity as existing precomputing methods, \emph{i.e.}, being linear with respect
to the number of nodes $n$. Also, in practice, the generic deep learning operations, such as convolution and attention, are well accelerated and parallelized in modern platforms. Hence, our method achieves better balance between performance and efficiency. We compare actual running time to demonstrate this in Section~\ref{sec:exp_time}.

\section{Discussions}\label{sec:discussion}

The main difference between graph and grid-like data lies in the
notion and properties of locality. Specifically, the numbers of
neighbors differ for different nodes, and there is no order
information among the neighbors of a node in graphs. These are the
main obstacles preventing the use of generic deep learning
operations on graphs. Our Neighbor2Seq is an attempt to bridge the
gap between graph and grid-like data. Base on our Neighbor2Seq, many
effective strategies for grid-like data can be naturally transferred
to graph data. These include self-supervised learning and
pre-training on
graphs~\cite{hu2019strategies,velickovic2019deep,sun2019infograph,hassani2020contrastive,you2020does,hu2020gpt,qiu2020gcc,jin2020self}.

We notice an existing work AWE~\cite{pmlr-v80-ivanov18a} which also embed the information in graph as a sequence. In order to avoid confusion, we make a clarification about the fundamental and significant differences between AWE and our Neighbor2Seq. First, AWE produces a sequence embedding for the entire graph, while our Neighbor2Seq yields a sequence embedding for each node in the graph. Second, each element in the obtained sequence in AWE is the probability of an anonymous walk embedding. In our Neighbor2Seq, each feature vector in the obtained sequence for one node is computed by summing up the features of all nodes in the corresponding layer of the neighborhood tree. This point distinguishes these two methods fundamentally.

More discussions about our method are included in the Supplementary Material.

\section{Experimental Studies}\label{sec:exp}

\subsection{Experimental Setup.}\label{sec:exp_setup}
\textit{(i) Datasets.} We evaluate our proposed models by performing node classification tasks on $1$
massive-scale graph (\textit{ogbn-papers100M}~\cite{hu2020open}) and $4$ medium-scale graphs (\textit{ogbn-products}~\cite{hu2020open},
\textit{Reddit}~\cite{hamilton2017inductive},
\textit{Yelp}~\cite{zeng2020graphsaint}, and
\textit{Flickr}~\cite{zeng2020graphsaint}). These tasks cover
inductive and transductive settings. The statistics of these
datasets are summarized in Table~\ref{tab:datsets}. The detailed
description of these datasets are provided in the Supplementary Material.

\textit{(ii) Implementation.} We implemented our methods using
Pytorch~\cite{paszke2017automatic} and Pytorch
Geometric~\cite{Fey2019pyg}. For our proposed methods, we conduct
the precomputation on a CPU, after which we train our models on a
GeForce RTX 2080 Ti GPU. We perform a grid search for the following
hyperparameters: \textit{number of hops $L$}, \textit{batch size},
\textit{learning rate}, \textit{hidden dimension $d'$}, \textit{dropout
    rate}, \textit{weight decay}, and \textit{convolutional kernel
    size $k$}. 

\subsection{Results on Massive-Scale Graphs.}
Since \textit{ogbn-papers100M} is a massive graph with more than
$111$ million nodes and $1.6$ billion edges, most existing methods
have difficulty handling such a graph. We consider three baselines
that have available results evaluated by OGB~\cite{hu2020open}: Multilayer Perceptron
(MLP), Node2Vec~\cite{grover2016node2vec}, and
SGC~\cite{wu2019simplifying}. Note that the existing sampling methods cannot be applied to this massive graph. The results under transductive
setting is reported in Table~\ref{tab:result_papers100M}. Following
OGB, we report accuracies for all models on training, validation,
and test sets. Our models outperform the baselines consistently in
terms of training, validation, and test. In particular, our Neighbor2Seq+Conv performs better than SGC by an obvious margin of $2.33\%$ in terms of test accuracy. This demonstrates the
expressive power and the generalization ability of our method on
massive graphs.

\subsection{Results on Medium-Scale Graphs.}

We also evaluate our models on medium-scale graphs, thus enabling comparison with more
existing strong baselines. We conduct
transductive learning on \textit{ogbn-products}, a medium-scale
graph from OGB. We also conduct inductive learning on
\textit{Reddit}, \textit{Yelp}, and \textit{Flickr}, which are
frequently used for inductive learning by the community. The
following various baselines are considered: MLP,
Node2Vec~\cite{grover2016node2vec}, GCN~\cite{kipf2016semi}, SGC~\cite{wu2019simplifying}, GraphSAGE~\cite{hamilton2017inductive},
FastGCN~\cite{chen2018fastgcn}, VR-GCN~\cite{chen2018stochastic},
AS-GCN~\cite{huang2018adaptive},
ClusterGCN~\cite{chiang2019cluster},
GraphSAINT~\cite{zeng2020graphsaint}, and
SIGN~\cite{rossi2020sign}.

\begin{table}[t]

	\caption{Results on \textit{ogbn-products} in terms of classification accuracy (in percent). The reported accuracy is averaged over $10$ random runs. Obtaining the results of GCN requires a GPU with 33GB of memory. The best performance on test set is highlighted in bold and the second performance is shown with an underline.}
	\label{tab:result_products}
	\centering
	\vspace{0.05in}
	\resizebox{1\columnwidth}{!}{
		\begin{tabular}{lccc}
			\toprule
			Method & Training & Validation & Test\\
			\midrule
			MLP & $84.03 \scriptstyle{{\light{\pm0.93}}}$ & $75.54\scriptstyle{{\light{\pm0.14}}}$ & $61.06\scriptstyle{{\light{\pm0.08}}}$ \\
			Node2vec & $93.39 \scriptstyle{{\light{\pm0.10}}}$ & $90.32\scriptstyle{{\light{\pm0.06}}}$ & $72.49\scriptstyle{{\light{\pm0.10}}}$ \\
			GCN & $93.56\scriptstyle{{\light{\pm0.09}}}$ & $92.00\scriptstyle{{\light{\pm0.03}}}$ & $75.64\scriptstyle{{\light{\pm0.21}}}$ \\
			GraphSAGE & $92.96\scriptstyle{{\light{\pm0.07}}}$ & $91.70\scriptstyle{{\light{\pm0.09}}}$ & $78.70\scriptstyle{{\light{\pm0.36}}}$ \\
			ClusterGCN & $93.75\scriptstyle{{\light{\pm0.13}}}$ & $92.12\scriptstyle{{\light{\pm0.09}}}$ & $78.97\scriptstyle{{\light{\pm0.33}}}$ \\
			GraphSAINT & $92.71\scriptstyle{{\light{\pm0.14}}}$ & $91.62\scriptstyle{{\light{\pm0.08}}}$ & $79.08\scriptstyle{{\light{\pm0.24}}}$ \\
			SGC & $92.60\scriptstyle{{\light{\pm0.10}}}$ & $91.19\scriptstyle{{\light{\pm0.06}}}$ & $72.46\scriptstyle{{\light{\pm0.27}}}$ \\
			SIGN & $96.92\scriptstyle{{\light{\pm0.46}}}$ & $93.10\scriptstyle{{\light{\pm0.08}}}$ & $77.60\scriptstyle{{\light{\pm0.13}}}$ \\
			\midrule
			\textbf{Neighbor2Seq+Conv} & $95.32\scriptstyle{{\light{\pm0.10}}}$ & $92.92\scriptstyle{{\light{\pm0.05}}}$ & $\textbf{79.67}\scriptstyle{{\light{\pm0.16}}}$ \\
			\textbf{Neighbor2Seq+Attn} & $92.82\scriptstyle{{\light{\pm0.14}}}$ & $92.20\scriptstyle{{\light{\pm0.02}}}$ & $\underline{79.35}\scriptstyle{{\light{\pm0.17}}}$ \\
			\bottomrule
		\end{tabular}
		}
\end{table}

The \textit{ogbn-products} dataset is challenging because the
splitting is not random. As described in Section~\ref{sec:exp_setup}, the splitting procedure is more realistic and matches the
real-world application where manual labeling is prioritized to
important nodes and models are subsequently used to make prediction
on less important nodes. Also, we only have $10\%$ nodes for training and validation in \textit{ogbn-products}. Hence, \textit{ogbn-products} is an ideal
benchmark dataset to evaluate the capability of out-of-distribution prediction. As
shown in Table~\ref{tab:result_products}, our Neighbor2Seq+Conv and
Neighbor2Seq+Attn outperfom baselines on test set (\emph{i.e.}, $90\%$
nodes), which further demonstrates the generalization ability of our
method on limited training data.

\begin{table}[t]
	\caption{Results for inductive learning on three datasets in terms of F1-micro score. The reported score is averaged over $10$ random runs. The results of baselines are partially obtained from~\cite{zeng2020graphsaint, rossi2020sign}. For each dataset, the best performance on test set is highlighted in bold and the second performance is shown with an underline.}
	\label{tab:result_inductive}
	\centering
	\vspace{0.05in}
	\resizebox{1\columnwidth}{!}{
		\begin{tabular}{lccc}
			\toprule
			Method & \textit{Reddit} & \textit{Flickr} & \textit{Yelp}\\
			\midrule
			GCN & $0.933 \scriptstyle{{\light{\pm0.000}}}$ & $0.492\scriptstyle{{\light{\pm0.003}}}$ & $0.378\scriptstyle{{\light{\pm0.001}}}$ \\
			FastGCN & $0.924 \scriptstyle{{\light{\pm0.001}}}$ & $0.504\scriptstyle{{\light{\pm0.001}}}$ & $0.265\scriptstyle{{\light{\pm0.053}}}$ \\
			VR-GCN & $0.964 \scriptstyle{{\light{\pm0.001}}}$ & $0.482\scriptstyle{{\light{\pm0.003}}}$ & $0.640\scriptstyle{{\light{\pm0.002}}}$ \\
			AS-GCN & $0.958 \scriptstyle{{\light{\pm0.001}}}$ & $0.504\scriptstyle{{\light{\pm0.002}}}$ & - \\
			GraphSAGE & $0.953 \scriptstyle{{\light{\pm0.001}}}$ & $0.501\scriptstyle{{\light{\pm0.013}}}$ & $0.634\scriptstyle{{\light{\pm0.006}}}$ \\
			ClusterGCN & $0.954 \scriptstyle{{\light{\pm0.001}}}$ & $0.481\scriptstyle{{\light{\pm0.005}}}$ & $0.609\scriptstyle{{\light{\pm0.005}}}$ \\
			GraphSAINT & $0.966 \scriptstyle{{\light{\pm0.001}}}$ & $0.511\scriptstyle{{\light{\pm0.001}}}$ & $\textbf{0.653}\scriptstyle{{\light{\pm0.003}}}$ \\
			SGC & $0.949 \scriptstyle{{\light{\pm0.000}}}$ & $0.502\scriptstyle{{\light{\pm0.001}}}$ & $0.358\scriptstyle{{\light{\pm0.006}}}$ \\
			SIGN & $\textbf{0.968} \scriptstyle{{\light{\pm0.000}}}$ & $0.514\scriptstyle{{\light{\pm0.001}}}$ & $0.631\scriptstyle{{\light{\pm0.003}}}$ \\
			\midrule
			\textbf{Neighbor2Seq+Conv} & $\underline{0.967}\scriptstyle{{\light{\pm0.000}}}$ & $\textbf{0.527}\scriptstyle{{\light{\pm0.003}}}$ & $\underline{0.647}\scriptstyle{{\light{\pm0.003}}}$ \\
			\textbf{Neighbor2Seq+Attn} & $\underline{0.967}\scriptstyle{{\light{\pm0.000}}}$ & $\underline{0.523}\scriptstyle{{\light{\pm0.002}}}$ & $\underline{0.647}\scriptstyle{{\light{\pm0.001}}}$ \\
			\bottomrule
		\end{tabular}
	}
\end{table}

The results on inductive tasks are summarized in
Table~\ref{tab:result_inductive}. On \textit{Reddit}, our models
perform better than all sampling methods and achieve the
competitive result as SIGN. On \textit{Flickr}, our models obtain
significantly better results. Specifically, our Neighbor2Seq+Conv
outperforms the previous state-of-the-art models by an obvious
margin. Although our models perform not as good as GraphSAINT on
\textit{Yelp}, we outperform other sampling methods and the
precomputing model SIGN consistently on this dataset. To the best of our knowledge, there does not exist a method that can perform best overwhelmingly on all these datasets for inductive tasks currently. Therefore, we believe the experimental results can clearly demonstrate the effectiveness of our methods for inductive learning.

\begin{table}[t]
	\caption{Computational efficiency in terms of preprocessing and training (per epoch) (in seconds) on \textit{ogbn-products}. The reported time is averaged over $10$ runs. The test performance is included for reference.}
	\label{tab:time_comp}
	\centering
	\vspace{0.05in}
		\resizebox{\columnwidth}{!}{
			\begin{tabular}{lcccc}
				\toprule
				Method & Preproc. ($\downarrow$) & Train ($\downarrow$)& Test Acc. ($\uparrow$) \\
				\midrule
				ClusterGCN & $44.15\scriptstyle{{\light{\pm0.77}}}$ & $11.87\scriptstyle{{\light{\pm0.84}}}$ &  $78.97\scriptstyle{{\light{\pm0.33}}}$  \\
				GraphSAINT & $80.78\scriptstyle{{\light{\pm3.5}}}$ & $4.29\scriptstyle{{\light{\pm0.48}}}$ &  $79.08\scriptstyle{{\light{\pm0.24}}}$  \\
				SGC & $153.36\scriptstyle{{\light{\pm3.6}}}$ & $0.15\scriptstyle{{\light{\pm0.01}}}$ &  $72.46\scriptstyle{{\light{\pm0.27}}}$  \\
				SIGN & $151.47\scriptstyle{{\light{\pm3.5}}}$ & $1.22\scriptstyle{{\light{\pm0.02}}}$ &  $77.60\scriptstyle{{\light{\pm0.13}}}$  \\
				\midrule
				\textbf{Neighbor2Seq+Conv} & $153.42\scriptstyle{{\light{\pm3.2}}}$ & $4.09\scriptstyle{{\light{\pm0.12}}}$ &  $79.67\scriptstyle{{\light{\pm0.16}}}$  \\
				\textbf{Neighbor2Seq+Attn} & $153.42\scriptstyle{{\light{\pm3.2}}}$ & $2.67\scriptstyle{{\light{\pm0.08}}}$ &  $79.35\scriptstyle{{\light{\pm0.17}}}$  \\
				\bottomrule
			\end{tabular}
		}
\end{table}

\begin{table*}[t]
    \caption{Comparison of models with and without capturing order information. Neighbor2Seq+Attn w/o PE denotes the Neighbor2Seq+Attn without adding positional encoding. For each dataset, the best performance on test set is highlighted in bold and the second performance is shown with an underline.}
    \label{tab:order_info}
    \centering
    \vspace{0.05in}
        \resizebox{2\columnwidth}{!}{
            \begin{tabular}{lcccccc}
                \toprule
                Model & Order information & \textit{ogbn-papers100M}  & \textit{ogbn-products} & \textit{Reddit} & \textit{Flickr} & \textit{Yelp} \\
                \midrule
                \textbf{Neighbor2Seq+Conv} & \cmark & $\textbf{65.62}\scriptstyle{{\light{\pm0.32}}}$ & $\textbf{79.67}\scriptstyle{{\light{\pm0.16}}}$ & $\textbf{0.967}\scriptstyle{{\light{\pm0.000}}}$ & $\textbf{0.527}\scriptstyle{{\light{\pm0.003}}}$ & $\textbf{0.647}\scriptstyle{{\light{\pm0.003}}}$ \\
                \textbf{Neighbor2Seq+Attn} & \cmark & $\underline{65.04}\scriptstyle{{\light{\pm0.24}}}$ & $\underline{79.35}\scriptstyle{{\light{\pm0.17}}}$ & $\textbf{0.967}\scriptstyle{{\light{\pm0.000}}}$ & $\underline{0.523}\scriptstyle{{\light{\pm0.002}}}$ & $\textbf{0.647}\scriptstyle{{\light{\pm0.001}}}$ \\
                \textbf{Neighbor2Seq+Attn w/o PE} & \xmark & $65.03\scriptstyle{{\light{\pm0.42}}}$ & $78.54\scriptstyle{{\light{\pm0.25}}}$ & $0.965\scriptstyle{{\light{\pm0.000}}}$ & $0.521\scriptstyle{{\light{\pm0.003}}}$ & $0.646\scriptstyle{{\light{\pm0.001}}}$ \\
                \bottomrule
            \end{tabular}
        }
\end{table*}

\subsection{Comparisons of Computional Efficiency.}\label{sec:exp_time}

In order to verify our analysis of time complexity in Section~\ref{sec:scalablitiy}, we conduct an empirical comparison with existing methods in terms of real running time during preprocessing and training. We consider the following representative sampling methods and precomputing methods: ClusterGCN~\cite{chiang2019cluster}, GraphSAINT~\cite{zeng2020graphsaint}, SGC~\cite{wu2019simplifying}, and SIGN~\cite{rossi2020sign}. The comparison is performed on \textit{ogbn-products} and the similar trend can be observed on other datasets. As demonstrated in Table~\ref{tab:time_comp}, our approaches, like existing precomputing methods, are more computationally efficient than sampling methods in terms of training. Although the precomputing methods cost more time on preprocessing, this precomputing only need to be conducted by one time. Compared with existing precomputing methods, our methods obtain better performance with introducing affordable computation, achieving a better balance between performance and efficiency.

\subsection{Ablation Study on Order Information}

Intuitively, the order information in the sequence obtained by
Neighbor2Seq indicates the hops between nodes. Hence, we conduct an
ablation study to verify the significance of this order information.
We remove the positional encoding in Neighbor2Seq+Attn, leading to a
model without the ability to capture the order information. The
comparison is demonstrated in Table~\ref{tab:order_info}. Note that
Neighbor2Seq+Attn and Neighbor2Seq+Attn w/o PE have the same number
of parameters. Hence, Comparing the results of these two models, we
can conclude that the order information is usually necessary and the degree of importance depends on the corresponding dataset.
Both Neighbor2Seq+Conv and Neighbor2Seq+Attn can capture the order
information. We observe that Neighbor2Seq+Conv
performs better. The possible reason could be that Neighbor2Seq+Conv has more learnable
parameters than Neighbor2Se+Attn, which only has a learnable query.

\section{Conclusions}

In this work, we propose Neighbor2Seq, for transforming the
heirarchical neighborhoods to ordered sequences. Neighbor2Seq
enables the subsequent use of powerful general deep learning
operations, leading to the proposed Neighbor2Seq+Conv and
Neighbor2Seq+Attn. Our models can be deployed on massive graphs and
trained efficiently. The extensive expriments demonstrate the
scalability and the promising performance of our method.

 \section*{Acknowledgment}

This work was supported in part by National Science Foundation grants IIS-2006861 and DBI-1922969.

\bibliographystyle{abbrv}
\bibliography{my_reference}




\end{document}